%% file: root.tex
\documentclass[conference]{IEEEtran}
\IEEEoverridecommandlockouts

\usepackage[left=54pt, right=54pt, bottom=54pt, top=54pt]{geometry}
\usepackage[pdftex]{graphicx}
\usepackage{booktabs}
\usepackage{moreverb}
\usepackage{url}
\usepackage{amsmath}
\usepackage{bm}
\usepackage[caption=false]{subfig}
\usepackage{color}
\usepackage{amsfonts}
\usepackage{amssymb}
\usepackage{multicol}
\usepackage{mathtools}
\usepackage[usenames,dvipsnames]{xcolor}
\usepackage{algpseudocode}
\usepackage{algorithm}
\usepackage{enumitem}
\usepackage[colorlinks,linkcolor=black,bookmarksopen,bookmarksnumbered,citecolor=black,urlcolor=black]{hyperref}
\usepackage{cleveref}
\usepackage{siunitx}
\usepackage{glossaries-extra}
\setabbreviationstyle[acronym]{long-short}
\glssetcategoryattribute{acronym}{nohyperfirst}{true}

\newacronym{frbd}{FRBD}{full rigid body dynamics}
\newacronym{fbd}{FBD}{full-body dynamics}
\newacronym{cd}{CD}{centroidal dynamics}
\newacronym{srbd}{SRBD}{single rigid body dynamics}
\newacronym{mpc}{MPC}{model predictive control}
\newacronym{ddp}{DDP}{differential dynamic programming}
\newacronym{oc}{OC}{optimal control}

\newcommand{\eref}[1]{Eq.~(\ref{#1})}
\newcommand{\fref}[1]{Fig.~\ref{#1}}
\newcommand{\sref}[1]{Section~\ref{#1}}
\DeclareMathOperator*{\atan}{atan}
\DeclareMathOperator*{\atantwo}{atan2}

\def\BibTeX{{\rm B\kern-.05em{\sc i\kern-.025em b}\kern-.08em
    T\kern-.1667em\lower.7ex\hbox{E}\kern-.125emX}}

\newlength{\tempdima}
\newcommand{\rowname}[1]{\rotatebox{0}{\makebox[\tempdima][c]{(\footnotesize #1)}}}
\newcommand{\orcid}[1]{\href{https://orcid.org/#1}{\includegraphics[width=0.6em]{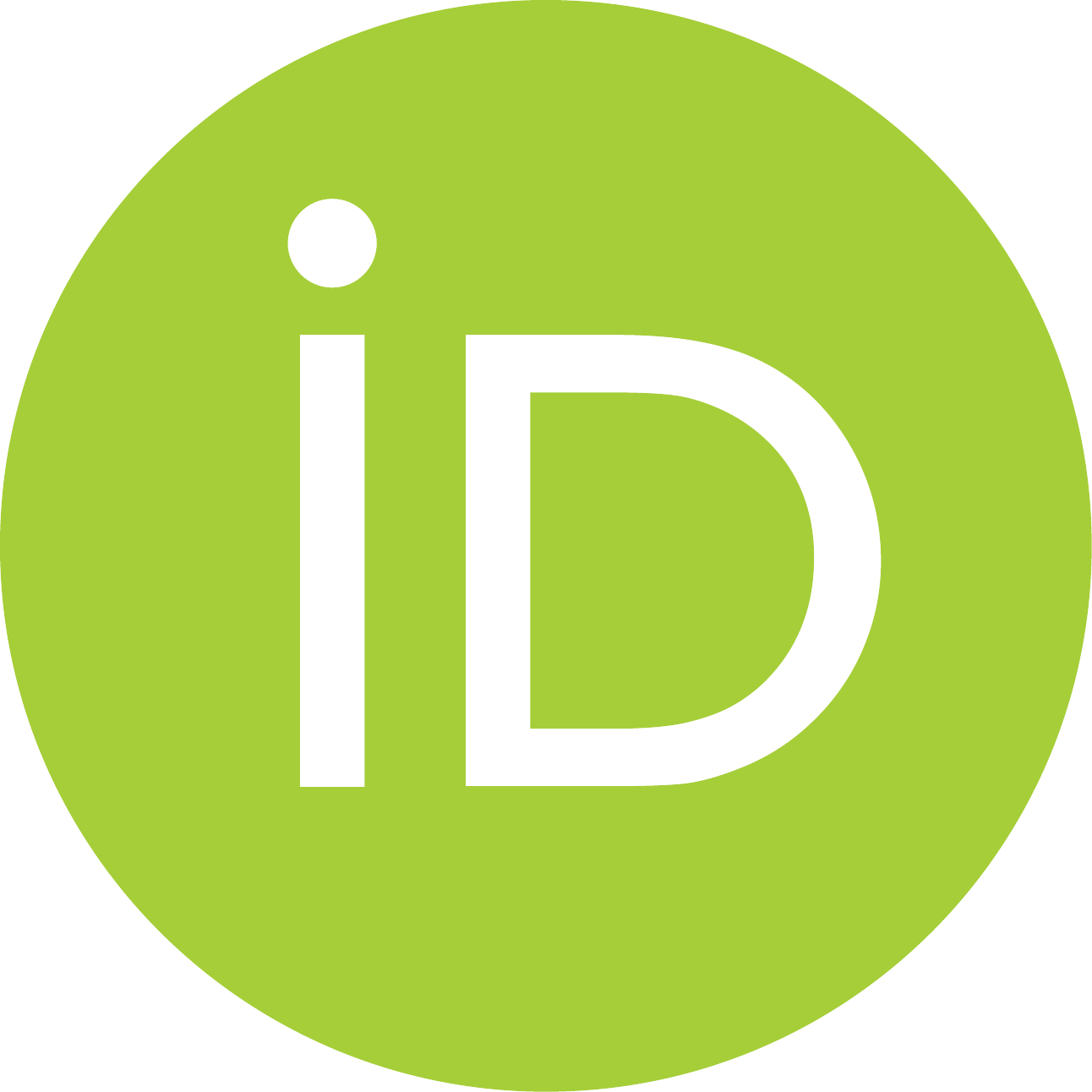}}}

\title{

\vspace*{18pt}Topology-Based MPC for Automatic\\
Footstep Placement and Contact Surface Selection}

\author{
    Jaehyun Shim\orcid{0000-0002-8008-7120}\quad
    Carlos Mastalli\orcid{0000-0002-0725-4279}\quad
    Thomas Corbères\orcid{0000-0002-3829-0096}\\
    \quad\quad Steve Tonneau\orcid{0000-0003-3001-8693}\quad
    Vladimir Ivan\orcid{0000-0002-6610-385X}\quad
    Sethu Vijayakumar\orcid{0000-0003-0649-7241}
\thanks{
Jaehyun Shim, Thomas Corbères, Steve Tonneau, and Sethu Vijayakumar are affiliated with the School of Informatics, University of Edinburgh, U.K. (e-mail: 
\href{mailto:jshim.robotics@gmail.com}{jshim.robotics@gmail.com},
\href{mailto:t.corberes@sms.ed.ac.uk}{t.corberes@sms.ed.ac.uk},
\href{mailto:stonneau@exseed.ed.ac.uk}{stonneau@exseed.ed.ac.uk},
\href{mailto:sethu.vijayakumar@ed.ac.uk}{sethu.vijayakumar@ed.ac.uk}).}
\thanks{
Carlos Mastalli is affiliated with the Institute of Sensors, Signals and Systems, School of Engineering and
Physical Sciences, Heriot-Watt University, U.K. (e-mail: \href{mailto:c.mastalli@hw.ac.uk}{c.mastalli@hw.ac.uk}).}
\thanks{
Vladimir Ivan is affiliated with Touchlab Limited, U.K. (e-mail: \href{mailto:v.ivan.mail@gmail.com}{v.ivan.mail@gmail.com}).

This research was conducted as part of the Memory of Motion (Memmo) project, a collaborative project supported by the European Union within the H2020 Program (Grant Agreement No. 780684).
}
}
\begin{document}

\maketitle

% ------------------------------ Abstract 
\input{src/0_abstract}

% ------------------------------ Introduction
\input{src/1_introduction}

% ------------------------------ Contact-region penalty
\input{src/2_contact_region_penalty}

% ------------------------------ MPC
\input{src/3_mpc}

% ------------------------------ Experiment
\input{src/4_results}

% ------------------------------ Conclusion
\input{src/5_conclusion}

% ------------------------------ Acknowledgment
% \input{src/6_acknowledgment}

% ------------------------------ Reference
\bibliography{reference}

\end{document}

%% file: src/0_abstract.tex
\begin{abstract}
State-of-the-art approaches to footstep planning assume reduced-order dynamics when solving the combinatorial problem of selecting contact surfaces in real time.
However, in exchange for computational efficiency, these approaches ignore joint torque limits and limb dynamics.
In this work, we address these limitations by presenting a topology-based approach that enables~\gls{mpc} to simultaneously plan full-body motions, torque commands, footstep placements, and contact surfaces in real time.
To determine if a robot's foot is inside a contact surface, we borrow the winding number concept from topology.
We then use this winding number and potential field to create a contact-surface penalty function.
By using this penalty function,~\gls{mpc} can select a contact surface from all candidate surfaces in the vicinity and determine footstep placements within it.
We demonstrate the benefits of our approach by showing the impact of considering full-body dynamics, which includes joint torque limits and limb dynamics, on the selection of footstep placements and contact surfaces.
Furthermore, we validate the feasibility of deploying our topology-based approach in an~\gls{mpc} scheme and explore its potential capabilities through a series of experimental and simulation trials.
\end{abstract}

%% file: src/1_introduction.tex
\section{Introduction}\label{sec:introduction}
To \textit{traverse discrete terrains} such as stepping stones, legged robots need to carefully plan their footsteps and motions ~\cite{kuindersma-ar16,mastalli-tro20,jenelten-tro22}.
In previous works, footsteps and motions were computed separately~\cite{winkler15icra,mastalli-icra20,tonneau18tro} to reduce the combinatorial complexity of these nonlinear problems~\cite{posa14ijrr,mastalli16icra}.
However, in doing so, assumptions need to be introduced into the gait pattern, kinematics, and dynamics model.
Alternatively, to focus on the footstep planning (i.e., combinatorial problem), other approaches neglect limb dynamics, thereby allowing for formulating this problem as a mixed-integer convex problem~\cite{aceituno_cabezas-ral18,tonneau-icra20}.
However, for robots with heavy limbs or limited actuation torque, this assumption does not hold, nor can the full reachability (i.e., kinematics) of the robot be exploited.
These limitations also lead to errors in footstep tracking, which are caused by improper tracking of angular momentum~\cite{mastalli-underreview22}.
The above-mentioned limitations can be addressed by considering the robot's full-body dynamics, as it also takes into account limb dynamics and joint torque limits, and computing motions and footsteps together.
However, this results in a large optimization problem that is challenging to solve within a control loop of a few milliseconds.

\begin{figure}[t]
    \centering
    \includegraphics[width=\linewidth]{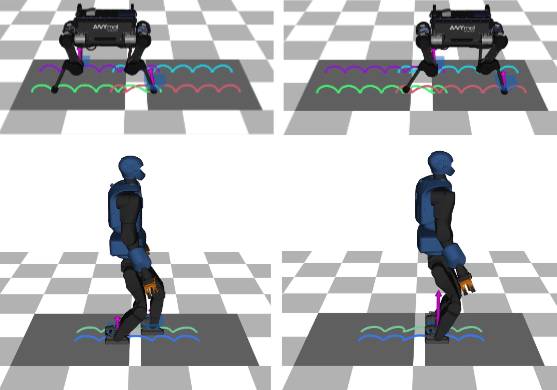}
    \caption{
    Visualization of the planned motions and footsteps, taking into account the robot's full-body dynamics (including joint torque limits and limb dynamics), friction cones, and all potential contact surfaces.
    Swing-foot trajectories are represented in different colors, while candidate contact surfaces are indicated by dark gray squares.
    Our approach can be used to plan footstep placements and contact surfaces for both quadruped and humanoid robots.
    A video demonstrating our approach is available at \texttt{\url{https://youtu.be/uweesAj5_x0}}.}
    \label{fig:footstep_planning}
\end{figure}

In our recent works, we demonstrated an~\gls{mpc} that takes into account the full-body dynamics of the robot~\cite{mastalli-underreview22,mastalli-invdynmpc22}.
A key advantage of our previous approaches is the ability of our~\gls{mpc} to generate agile and complex maneuvers through the use of feasibility-driven search~\cite{mastalli22auro} and optimal policy tracking.
However, it requires a predefined sequence of footstep placements and contact surfaces.
To address these limitations, this paper borrows concepts from topology and classical electrostatics to enable the automatic selection of footstep placements and contact surfaces in real time.
Compared to other state-of-the-art approaches, our footstep plans ensure joint torque limits, friction-cone constraints, and full-body kinematics and dynamics (\fref{fig:footstep_planning}).
To the best of our knowledge, our work is the first to \textit{introduce full-body dynamics~\gls{mpc} that optimizes footstep placement and contact surface} as well.

\subsection{Related Work}\label{sec:related_work}
Recent methods for footstep planning can optimize footstep placement and gait pattern~\cite{aceituno_cabezas-ral18, dai-ichr14, winkler-ral18, jiayi20iros}.
These approaches can handle discrete terrains by smoothing their geometry~\cite{posa14ijrr, dai-ichr14, winkler-ral18} or using discrete variables in mixed-integer optimization~\cite{aceituno_cabezas-ral18, tonneau-icra20, jiayi20iros, deits14ichr}.
However, their computational complexity makes it infeasible to deploy them online.
This is because they result in a combinatorial explosion of hybrid modes, which cannot be resolved by employing simplified models to cast the problem as mixed-integer convex optimization~\cite{aceituno_cabezas-ral18, deits14ichr}.
Alternatively, it can be formulated as a continuous problem with complementary constraints.
However, the ill-posed nature of the complementary constraints increases computation time, making it difficult to deploy this approach online~\cite{posa14ijrr,mastalli16icra,dai-ichr14}.
To avoid combinatorial complexity or ill-conditioning, our work \textit{focuses on optimizing footstep placement without optimizing gait pattern}.

The narrow focus on optimizing footstep placement enables online re-planning.
For instance, as shown in~\cite{herdt10ar}, \gls{mpc} can generate a walking motion with automatic footstep placement.
However, despite the impressive achievement, this approach assumes that the robot behaves as a linear inverted pendulum, which cannot account for its kinematic feasibility, joint torque limits, orientation, or the effect of non-coplanar contact conditions and limb and vertical motions.
Later, Di Carlo et al.~\cite{dicarlo-iros18} proposed a convex relaxation of the~\gls{srbd} that can accommodate the robot's orientation.
As in~\cite{herdt10ar}, this boils down to a linear~\gls{mpc} that can be solved with a general-purpose quadratic programming solver.
More recently, other~\gls{mpc} approaches employ~\gls{srbd} or~\gls{cd}, using direct transcription and general-purpose nonlinear programming solvers~\cite{bledt19iros,romualdi-icra22}.
Although these approaches can address non-coplanar contacts and vertical motions,~\gls{srbd} still cannot account for the robot's kinematic limits and the effects of limb dynamics.
Moreover, neither~\gls{srbd} nor~\gls{cd} can account for joint torque limits.
One may think that these limitations are not critical for robots with lightweight legs; however, a recent study shows that they still have a substantial impact on the control~\cite{corberes-icra21}.
This justifies why, in our work, we \textit{compute motions that ensure the robot's full-body dynamics}.

Real-time handling of both body and leg kinematics and dynamics in an~\gls{mpc} manner can be achieved by taking advantage of the temporal structure of the optimal control problem.
\Gls{ddp}~\cite{mayne-66} takes advantage of this structure by factorizing a sequence of smaller matrices, rather than employing sparse linear solvers~\cite{HSL} commonly done in nonlinear programming~\cite{nocedal-optbook}.
This reduction in computational complexity makes it feasible to use~\gls{mpc} with full-body dynamics, as shown in simulation results presented in~\cite{tassa-iros12}.
Inspired by these results, recent research has shown the application of~\gls{mpc} with~\gls{srbd} and full kinematics~\cite{farshidian-ichr17,grandia19iros,grandia21icra} and full-body dynamics~\cite{mastalli-underreview22,mastalli-invdynmpc22,koenemann-iros15,neunert-ral18,katayama-underreview}.
However, these approaches do not address the issue of selecting footstep placement and contact surface with full-body dynamics.
While algorithms based on~\gls{ddp}, such as iLQR~\cite{li-icinco04}, \textsc{Box-FDDP}~\cite{mastalli22auro}, and \textsc{ALTRO}~\cite{howell-19}, have been also developed, they have not yet been applied to address the aforementioned problems.
In contrast to other~\gls{mpc} approaches, our topology-based approach selects optimal footstep placement and contact surface within the framework of full-body dynamics~\gls{mpc} by \textit{creating a continuous cost function for terrain}.

\subsection{Contribution}
The main contribution of our work is an~\gls{mpc} that simultaneously plans full-body motions, torque commands, feedback policies, footstep placements, and contact surfaces in real time.
Specifically, we identify three technical contributions as follows:
\begin{enumerate}[label=(\roman*)]
    \item A novel topology-based~\gls{mpc} that uses potential field and winding number to plan footstep placements and contact surfaces in real time;
    \item Demonstration of the advantages of our topology-based~\gls{mpc} that takes into account full-body dynamics;
    \item Experimental validation of our topology-based~\gls{mpc} on the ANYmal robot and simulations that showcase its potential capabilities.
\end{enumerate}
In the next section, we introduce the concepts of potential field and winding number that we use to \textit{create a contact-surface penalty function}.

%% file: src/2_contact_region_penalty.tex
\section{Potential Field and Winding Number in Contact Surfaces}\label{sec:methodology}
In this section, we explain in detail our novel approach to footstep placement and contact surface selection.
It borrows the concepts of electric potential (\sref{sec:electric_potential} and \ref{sec:potential_regions}) and winding number (\sref{sec:winding_number} and \ref{sec:winding_number_regions}) from classical electrostatics and topology, respectively.
These concepts enable us to design a contact-surface penalty function, allowing our~\gls{mpc} to choose optimal footstep placement and contact surface (\sref{sec:harmonic_cost}).

\subsection{Potential Field}\label{sec:electric_potential}
We begin with an introduction to the electric potential that arises from a charged particle.
This is also called a Coulomb potential and is defined as
\begin{equation}
    V_E(\mathbf{r}) = \frac{Q}{4\pi\varepsilon_0}\frac{1}{\left|\mathbf{r} - \mathbf{p}\right|},
\end{equation}
where $Q$ is the charge of the particle, $\varepsilon_0$ is the permittivity of vacuum, $\mathbf{r}$ is the point at which the potential is evaluated, and $\mathbf{p}$ is the point at which the charged particle is located.
Since the physical meaning of the electric potential is not relevant to the definition of our cost function, we drop the scaling factor $\frac{Q}{4\pi\varepsilon_0}$ and make the potential unitless.
Moreover, the potential measured across a closed curve $\boldsymbol{\gamma}(s)$ can be obtained by integrating the potential field:
\begin{equation}\label{eq:potential}
    V(\boldsymbol{\gamma}) = \int\frac{1}{\left|\boldsymbol{\gamma}(s) - \mathbf{p}\right|} ds.
\end{equation}

\subsection{Potential Field in Contact Surfaces}\label{sec:potential_regions}
We describe the contact surfaces as polygons, whose boundaries can be represented by a sequence of linear segments.
Similar to~\eref{eq:potential}, we calculate the potential measured across the $i$-th segment of the polygon as follows:
\begin{equation}\label{eq:potential_segment}
    p_i = \int_0^1\frac{1}{\left|\mathbf{a}_i + (\mathbf{b}_i-\mathbf{a}_i)s - \mathbf{\hat{p}}\right|} ds,
\end{equation}
where $\mathbf{a}_i, \mathbf{b}_i\in\mathbb{R}^2$ are the endpoints of the segment and $\mathbf{\hat{p}}\in\mathbb{R}^2$ is the position of the robot's foot in the \textit{horizontal plane}.
It has an analytical solution, and the resulting potential produced by each potential segment $p_i$ can be calculated as follows:
\begin{equation}\label{eq:potential_region}
    V_{\boldsymbol{\gamma}} = -\sum_i^{\boldsymbol{\gamma}}\frac{\atan(\frac{c_i}{e_i}) - \atan(\frac{d_i}{e_i})}{e_i},
\end{equation}
with\vspace{-1.5em}
\begin{align*}
c_i &= (\mathbf{a}_i\cdot\mathbf{b}_i) - (\mathbf{a}_i\cdot\mathbf{\hat{p}}) + (\mathbf{b}_i\cdot\mathbf{\hat{p}}) - (\mathbf{b}_i\cdot\mathbf{b}_i), \\
d_i &= - (\mathbf{a}_i\cdot\mathbf{b}_i) - (\mathbf{a}_i\cdot\mathbf{\hat{p}}) + (\mathbf{b}_i\cdot\mathbf{\hat{p}}) + (\mathbf{a}_i\cdot\mathbf{a}_i), \\
e_i &= [\mathbf{a}_i\times\mathbf{b}_i]_z - [\mathbf{a}_i\times\mathbf{\hat{p}}]_z + [\mathbf{b}_i\times\mathbf{\hat{p}}]_z,
\end{align*}
where $[\mathbf{a}\times \mathbf{b}]_z$ returns the $z$-coordinate of the cross product of the two vectors $\mathbf{a} = [a_x, a_y]$ and $\mathbf{b} = [b_x, b_y]$ in the XY plane (i.e., $a_x b_y - a_y b_x$).

\subsection{Winding Number}\label{sec:winding_number}
To represent discrete terrain as a topological space, we leverage the concept of \textit{winding number}.
The winding number is a measure of how many times a curve is wound around a point in the 2D plane.
There are different ways to define the winding number.
For instance, from the perspective of differential geometry, we can associate this number with the polar coordinates for a point at the origin ($\mathbf{\hat{p}=0}$), i.e.,
\begin{equation}
    \text{wind}(\boldsymbol{\gamma}, \mathbf{\hat{p}=0}) = \frac{1}{\pi}\oint_{\boldsymbol{\gamma}}\left(\frac{x}{r^2}dy + \frac{y}{r^2}dx\right),
\end{equation}
where $x, y$ defines the parametric equation of a continuous closed curve $\boldsymbol{\gamma}$, with $r^2 = x^2 + y^2$.

\subsection{Winding Number in Contact Surfaces}\label{sec:winding_number_regions}
As in~\sref{sec:potential_regions} and using the formula derived by~\cite{O'Rourke:1998:CGC:521378}, we calculate the winding number of the contact surface by summing over the $i$-th linear segments, as follows:

\begin{equation}\label{eq:winding}
    \text{wind}(\boldsymbol{\gamma}, \mathbf{\hat{p}}) = \sum_i^{\boldsymbol{\gamma}} \frac{\atantwo(c_i,d_i)}{2\pi},
\end{equation}
with\vspace{-1.5em}
\begin{align*}
    c_i &= [(\mathbf{a}_i-\mathbf{\hat{p}})\times(\mathbf{b}_i-\mathbf{\hat{p}})]_z, \\
    d_i &= (\mathbf{a}_i-\mathbf{\hat{p}})\cdot(\mathbf{b}_i-\mathbf{\hat{p}}). \nonumber 
\end{align*}
Then, we can determine if a point $\mathbf{\hat{p}}$ is inside the contact surface by checking if $\text{wind}(\boldsymbol{\gamma}, \mathbf{\hat{p}})\geq\frac{1}{2}$.

\subsection{Contact-Surface Penalty Function}\label{sec:harmonic_cost}
To enable the robot to select a placement $\mathbf{p}_{\mathcal{C}_k}$ (with $\mathcal{C}_k$ as each foot) within a contact surface, we assign a zero cost when the placement is inside the surface.
To achieve this, we use an indicator function based on the winding number to compute the \textit{contact-surface penalty} cost $\ell_{\mathbf{p}_{\mathcal{C}_k}}$ as follows:
\begin{align}\label{eq:indicator_function}
    \ell_{\boldsymbol{\mathbf{p}}_{\mathcal{C}_k}} = 
    \begin{cases}
        \quad 0 & \text{if } \sum_j V_{\boldsymbol{\gamma}_j} = 0, \\
        \sqrt{\frac{\sum_j N_j}{\sum_j V_{\boldsymbol{\gamma}_j}}} & \text{if } \sum_j \text{wind}(\boldsymbol{\gamma}_j,\mathbf{\hat{p}}_{\mathcal{C}_k})<\frac{1}{2}, \\
        \quad 0 & \text{if } \sum_j \text{wind}(\boldsymbol{\gamma}_j,\mathbf{\hat{p}}_{\mathcal{C}_k})\geq\frac{1}{2},
    \end{cases}
\end{align}
where $V_{\boldsymbol{\gamma}_j}$ is the potential, $N_j$ is the number of linear segments, and $\text{wind}(\boldsymbol{\gamma}_j,\mathbf{\hat{p}}_{\mathcal{C}_k})$ is the winding number obtained for the polygon $j$.
The potential and winding number are computed using~\eref{eq:potential_region} and~\eref{eq:winding}.
The analytical derivatives of this penalty function can be computed using the chain rule.

\begin{figure}[t]
    \centering
    \includegraphics[width=\linewidth]{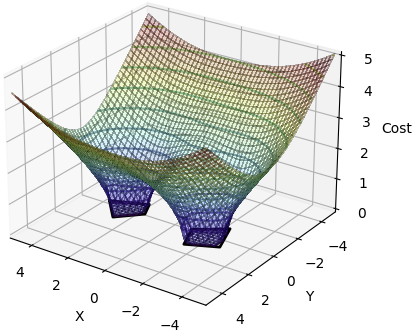}
    \caption{
    Contact-surface penalty function for two contact surfaces.
    The cost is zero inside the two candidate contact surfaces outlined in black and positive elsewhere.
    To determine if the robot's feet are inside the contact surfaces, we compute the winding number.
    }
    \label{fig:harmonic_cost}
\end{figure}

As shown in~\eref{eq:indicator_function}, we define the penalty function by using a square root (second line).
However, to avoid division by zero, we set the penalty function to zero when the potential is zero (first line).
This happens when $\mathbf{p}_{\mathcal{C}_k}$ lies on the boundary of the polygon. 
\fref{fig:harmonic_cost} illustrates our penalty function when we have two contact surfaces.

Our penalty function is based on electric potential, which results in a harmonic field that remains harmonic for convex polygons, non-convex polygons, collections of polygons, self-intersecting curves, and overlapping polygons.
This means that it contains the fewest possible number of local minima and saddle points.
Our topology-based representation allows us to enforce all of these properties using the winding number.
Furthermore, inspired by~\cite{ivan-ijrr2013}, our penalty function exploits invariances within a homology class defined by the winding number.
In other words, this function captures containment, which makes it an ideal candidate for numerical optimization.

%% file: src/3_mpc.tex
\section{MPC and Pipeline}
In this section, we describe our~\gls{mpc} formulation for selecting footstep placements and contact surfaces (\sref{sec:topology_based_mpc}).
Then, we present the control pipeline and setup used in our experiments with the ANYmal robot (\sref{sec:control_pipeline_and_experimental_setup}).

\subsection{Topology-Based MPC}\label{sec:topology_based_mpc}
Building upon our previous work~\cite{mastalli-underreview22}, our~\gls{mpc} solves a hybrid optimal control problem at each control time step.
The different modes of hybrid dynamics define different contact conditions along the optimization horizon.
These rigid contact conditions are subject to the robot's full-body dynamics (i.e., \textit{contact dynamics}).
We also model the contact-gain transitions between these modes using the \textit{impulse dynamics}.
Our~\textsc{Box-FDDP} solver~\cite{mastalli22auro} then computes full-body motions, torque commands, and feedback policies while keeping within the robot's joint torque limits, given a predefined set of footstep placements.

Here, our contact-surface penalty function extends the capabilities of our previous~\gls{mpc} by enabling it to automatically plan footstep placements and contact surfaces, given a reference velocity and a set of candidate contact surfaces.
In~\eref{eq:oc_problem}, the modifications introduced in this work are highlighted in \textcolor{blue}{blue}.

\begin{equation}\label{eq:oc_problem}
\resizebox{\columnwidth}{!}{$
\begin{aligned}
\min_{\mathbf{x}_s,\mathbf{u}_s}
&\hspace{-2.em}
& & \hspace{-0.75em}\sum_{k=0}^{N-1} \left(\ell^{reg}_k+\textcolor{blue}{ w_r\|\mathbf{\dot{r}}_k-\mathbf{\dot{r}}^{ref}\|^2 + w_f\sum_{\mathcal{C}_k}\ell_{\boldsymbol{\mathbf{p}}_{\mathcal{C}_k}}}\right) \hspace{-8.em}&\\
& \hspace{-1.5em}\textrm{s.t.} & &\hspace{-1em}\text{if $k$ is a contact-gain transition:}\\
& & & \mathbf{q}_{k+1} = \mathbf{q}_{k},\\
& & & \left[\begin{matrix}\mathbf{v}_{k+1} \\ -\boldsymbol{\Lambda}_{\mathcal{C}_k}\end{matrix}\right] =
\left[\begin{matrix}\mathbf{M}_k & \mathbf{J}^{\top}_{\mathcal{C}_k} \\ {\mathbf{J}_{\mathcal{C}_k}} & \mathbf{0} \end{matrix}\right]^{-1}
\left[\begin{matrix}\boldsymbol{\tau}^\mathcal{I}_{b_k} \\ -\mathbf{a}^\mathcal{I}_{\mathcal{C}_k} \\\end{matrix}\right], \hspace{-1em}&\textrm{(impulse dyn.)}\\
& & & \hspace{-1em}\textrm{else:}\\
& & & \mathbf{q}_{k+1} = \mathbf{q}_k \oplus \int_{t_k}^{t_k+\Delta t_k}\hspace{-2em}\mathbf{v}_{k+1}\,dt, &\\
& & & \mathbf{v}_{k+1} = \mathbf{v}_k + \int_{t_k}^{t_k+\Delta t_k}\hspace{-2em}\mathbf{\dot{v}}_k\,dt, &\textrm{(integrator)}\\
& & & \left[\begin{matrix}\mathbf{\dot{v}}_k \\ -\boldsymbol{\lambda}_{\mathcal{C}_k}\end{matrix}\right] =
\left[\begin{matrix}\mathbf{M}_k & \mathbf{J}^{\top}_{\mathcal{C}_k} \\ {\mathbf{J}_{\mathcal{C}_k}} & \mathbf{0} \end{matrix}\right]^{-1}
\left[\begin{matrix}\boldsymbol{\tau}^\mathcal{C}_{b_k} \\ -\mathbf{a}^\mathcal{C}_{\mathcal{C}_k} \\\end{matrix}\right], \hspace{-1em}&\textrm{(contact dyn.)}\\
  & & & \hspace{-1em}\textcolor{blue}{\log{({}^\mathcal{W}\mathbf{p}_{\mathcal{G}_k, z}^{-1}\cdot {}^\mathcal{W}\mathbf{p}^{ref}_{{\mathcal{G}_k, z}})} = \mathbf{0}}, &\textcolor{blue}{\textrm{(vertical foot pos.)}}\\
& & & \hspace{-1em}\textcolor{blue}{{}^\mathcal{W}\mathbf{\dot{p}}_{\mathcal{G}_k, z}^{-1}- {}^\mathcal{W}\mathbf{\dot{p}}^{ref}_{{\mathcal{G}_k, z}} = \mathbf{0}}, &\textcolor{blue}{\textrm{(vertical foot vel.)}}\\
% & & & \hspace{-1em}\log{({}^\mathcal{W}\mathbf{p}_{\mathcal{G}_k}^{-1}\cdot {}^\mathcal{W}\mathbf{p}^{ref}_{{\mathcal{G}_k}})} = \mathbf{0}, &\textrm{(contact pos.)}\\
% & & & \hspace{-1em}{}^\mathcal{W}\mathbf{\dot{p}}_{\mathcal{G}_k}^{-1}- {}^\mathcal{W}\mathbf{\dot{p}}^{ref}_{{\mathcal{G}_k}} = \mathbf{0}, &\textrm{(contact velocity)}\\
& & & \hspace{-1em}\textcolor{blue}{\mathbf{A}{}^\mathcal{W}\mathbf{p}^{ref}_{\mathcal{G}_k, z} = \mathbf{a}}, &\textcolor{blue}{\textrm{(contact surface height)}}\\
& & & \hspace{-1em}\mathbf{C}\boldsymbol{\lambda}_{\mathcal{C}_k} \geq \mathbf{c}, &\textrm{(friction-cone)}\\
& & & \hspace{-1em}\mathbf{\underline{x}} \leq \mathbf{x}_k \leq \mathbf{\bar{x}}, &\textrm{(state bounds)}\\
& & & \hspace{-1em}\mathbf{\underline{u}} \leq \mathbf{u}_k \leq \mathbf{\bar{u}}, &\textrm{(control bounds)}\\
\end{aligned}
$}
\end{equation}
where $\mathbf{x}=(\mathbf{q},\mathbf{v})$ is the state of the system, $\mathbf{q}\in\mathbb{SE}(3)\times\mathbb{R}^{n_j}$ (with $n_j$ as the number of joints) is the joint configuration, $\mathbf{v}\in\mathbb{R}^{n_v}$ (with $n_v = 6 + n_j$) is the generalized velocity, $\mathbf{u}\in\mathbb{R}^{n_j}$ is the joint torque input, $\boldsymbol{\lambda}_\mathcal{C}\in\mathbb{R}^{n_c}$ (with $n_c$ as the dimension of the contact forces) is the contact force, $\boldsymbol{\mathbf{p}}_{\mathcal{C,G}}\in\mathbb{R}^{n_c}$ (with
$\mathcal{C}$ and $\mathcal{G}$ as active and inactive contacts, respectively) is the position of the foot, $\mathbf{A}$ and $\mathbf{a}$ describes the contact surface, $\oplus$ and $\ominus$ denote \textit{integration} and \textit{difference operations} needed to optimize over manifolds~\cite{gabay82jota}---notations introduced in \textsc{Crocoddyl}~\cite{mastalli-icra20}.
Furthermore, $\ell^{reg}_k$ regularizes the robot's configuration around a nominal posture $\mathbf{q}^{ref}$, the generalized velocity, joint torques, and contact forces as follows:
\begin{equation*}
    \ell^{reg}_k = \|\mathbf{q}_k\ominus\mathbf{q}^{ref}\|^2_\mathbf{Q}+\|\mathbf{v}_k\|^2_\mathbf{N}+\|\mathbf{u}_k\|^2_\mathbf{R}+\|\boldsymbol{\lambda}_{\mathcal{C}_k}\|^2_\mathbf{K},
\end{equation*}
where $\mathbf{Q},\mathbf{N}\in\mathbb{R}^{n_v\times n_v}$, $\mathbf{R}\in\mathbb{R}^{n_j\times n_j}$ and $\mathbf{K}\in\mathbb{R}^{n_c\times n_c}$ are a set of (positive definite) diagonal weighting matrices.

To enable the automatic footstep placement, we first define a quadratic cost $w_r\|\mathbf{\dot{r}}_k-\mathbf{\dot{r}}^{ref}\|^2$ (with $w_r$ as its weight) that tracks the reference base velocity $\mathbf{\dot{r}}^{ref}$ in the horizontal plane.
Next, for all active contacts, we include our contact-surface penalty function $\sum_{\mathcal{C}_k}\ell_{\mathbf{p}_{\mathcal{C}_k}}$ (with $w_f$ as its weight).
We also add a constraint that specifies the contact surface height.
Finally, we impose constraints that define the vertical motion of the swing foot.
% Although it is possible to impose collision avoidance constraints, we choose to define constraints on the vertical movement of the foot so we can focus on validating our approach.
Note that we define the $\log(\cdot)$ operator in the vertical foot velocity constraint as the footstep placement may lie on a $\mathbb{SE}(3)$ manifold. ${}^\mathcal{W}\mathbf{p}_\mathcal{G}^{-1} \cdot {}^\mathcal{W}\mathbf{p}^{ref}_\mathcal{G}$ describes the inverse composition between the reference and current contact placements~\cite{blanco-10se3}.

\begin{figure}[t]
    \centering
    \includegraphics[trim={1.7cm 16.5cm 10.1cm 1.8cm}, clip, width=1.0\linewidth]{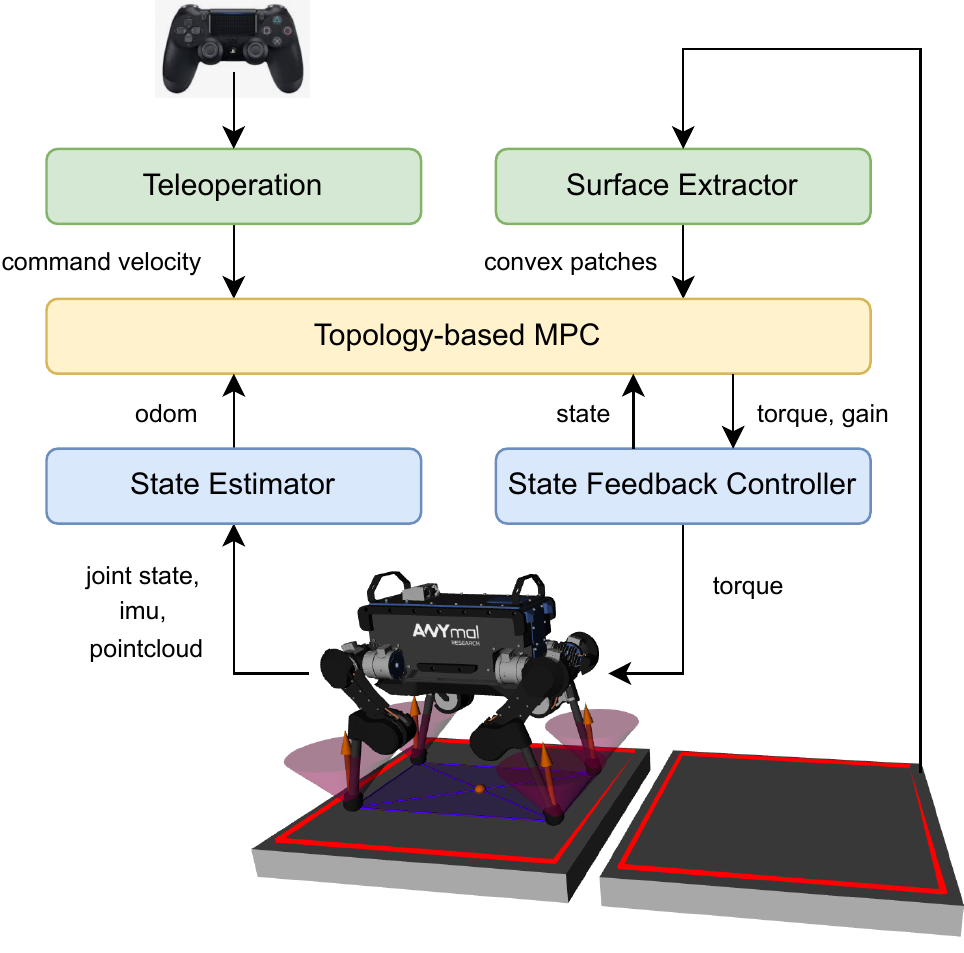}
    \caption{
    Overview of our locomotion pipeline.
    }
    \label{fig:control_pipeline}
\end{figure}

Again, we employ the robot's full-body dynamics (contact and impulse dynamics) in~\eref{eq:oc_problem}, where $\mathbf{M}\in\mathbb{R}^{n_v\times n_v}$ is the joint-space inertia matrix, $\mathbf{J}_\mathcal{C}\in\mathbb{R}^{nc \times nv}$ is the active contact Jacobian, $\boldsymbol{\tau}^{\mathcal{C},\mathcal{I}}_b\in\mathbb{R}^{n_v}$ is the force-bias vector, $\mathbf{a}_{\mathcal{C}}\in\mathbb{R}^{n_c}$ is the desired acceleration, and $\mathbf{C}$ and $\mathbf{c}$ describes the linearized friction cone (or wrench cone for humanoids).
Note that the definition of the force-bias term ($\boldsymbol{\tau}^{\mathcal{C}}_b$ or $\boldsymbol{\tau}^{\mathcal{I}}_b$) changes between the contact and impulse dynamics.
For more details on the regularization cost, dynamics, friction cone, and implementation aspects of our~\gls{mpc}, we encourage readers to refer to~\cite{mastalli-underreview22}.

\subsection{Locomotion Pipeline and Experimental Setup}\label{sec:control_pipeline_and_experimental_setup}

\fref{fig:control_pipeline} depicts our locomotion pipeline.
Reference velocities are sent using a joystick that runs at \SI{30}{\hertz}.
Candidate footstep surfaces in the predefined environment are extracted from mesh files at \SI{10}{\hertz}.
Our topology-based~\gls{mpc} operates at \SI{50}{\hertz} with an optimization horizon of \SI{0.85}{\second}.

Our~\gls{mpc}, teleoperation, and surface extractor run on two external PCs.
The joystick and surface extractor run on an Intel Core i9-9980 PC (8 core, 2.40 GHz), while the~\gls{mpc} runs on an Intel Core i9-9900 PC (8 core, 3.60 GHz).
Lastly, the state feedback controller and state estimator operate at \SI{400}{\hertz} on two separate onboard PCs equipped with an Intel Core i7-7500 (2 core, 2.70 GHz).

%% file: src/4_results.tex
\section{Results}\label{sec:results}
In this section, we first show the impact of considering the robot's full-body dynamics when selecting footstep placements and contact surfaces (\sref{sec:frbd_for_footstep_planning}).
This demonstrates the benefits of our approach compared to other state-of-the-art methods that use simplified models.
We then validate our approach in an~\gls{mpc} scheme on the ANYmal robot (\sref{sec:robot_experiment}).
Finally, we showcase the potential highly-dynamic maneuvers that our approach can generate by exploiting limb dynamics in simulations (\sref{sec:potential_dynamic_motion_with_high_torque}).

\begin{figure}[b!]
    \centering\begin{tabular}{cc}
    \rowname{a} & {\raisebox{-.5\height}{\includegraphics[width=0.93\columnwidth]{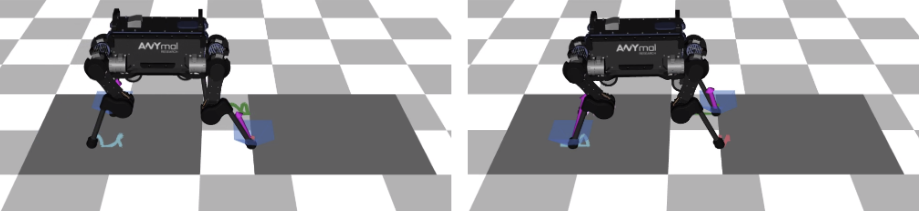}}}\\\\
    \rowname{b} & {\raisebox{-.5\height}{\includegraphics[width=0.93\columnwidth]{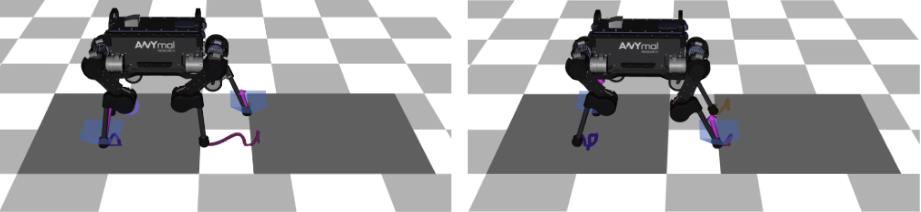}}}\\\\
    \rowname{c} & {\raisebox{-.5\height}{\includegraphics[width=0.93\columnwidth]{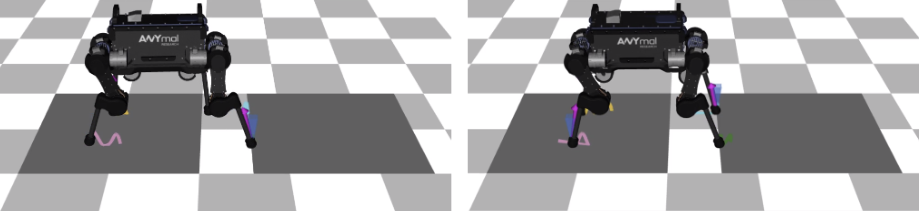}}}
    \end{tabular}
    \caption{
    Visualization of footstep placements and contact surfaces planned by our~\gls{mpc} when changing the joint torque limit and friction coefficient.
    The figures in the left and right columns of each row show the start and end of the crossing of the second foot, respectively.
    (a) ANYmal's default torque limits of \SI{40}{\newton\meter} and high friction coefficient of $1.0$.
    (b) Reduced torque limits of \SI{20}{\newton\meter}.
    (c) Low friction coefficient of $0.14$.
    When the torque limit is reduced in (b), our approach moves the footstep closer to the robot's hip and selects contact surfaces that require lower torque commands.
    Conversely, when the friction coefficient is low, our approach stretches the robot's leg perpendicular to the ground to maintain the reaction force within the small friction cone.
    }
    \label{fig:frbd_for_footstep_planning}
\end{figure}

\begin{figure}[!hbt]
    \centering
    \includegraphics[width=\linewidth]{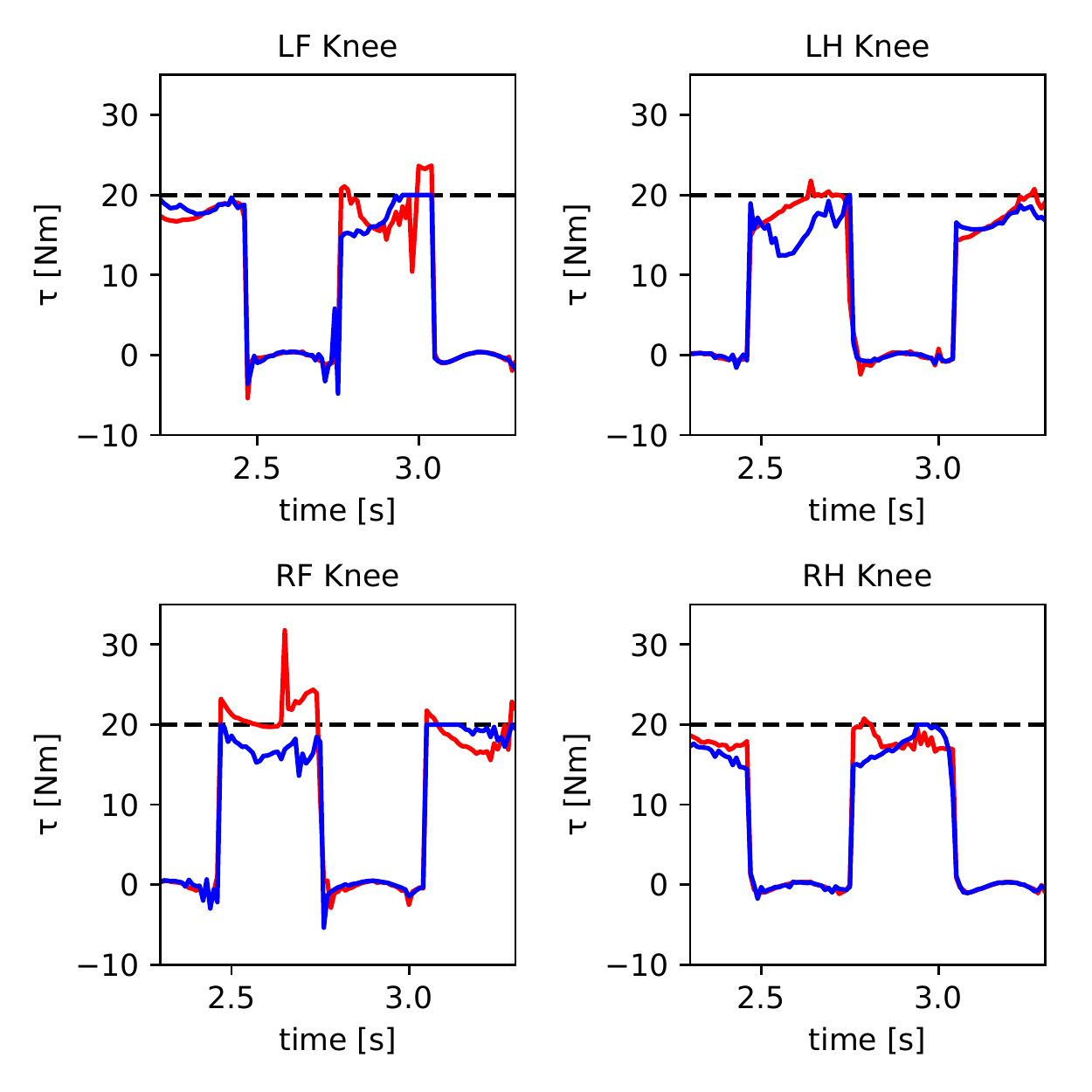}
    \caption{
    Torque commands on the knee joints when the robot crosses the patches. 
    The red line represents the torque commands computed with the ANYmal's default torque limit of \SI{40}{\newton\meter}, while the blue line represents the torque commands computed with the reduced torque limit of \SI{20}{\newton\meter}.
    The black dashed line indicates the reduced torque limit.
    With the default torque limit, the torque command for the right front knee computed by our~\gls{mpc} reaches its peak at~\SI{2.7}{\second}.
    When the torque limit is lowered, our~\gls{mpc} changes the patch crossing moment, as shown in \fref{fig:frbd_for_footstep_planning}b.
    The command torque for the left front knee reaches the limit at~\SI{2.9}{\second}, but it is maintained within it during that gait.
    }
    \label{fig:torque_plot}
\end{figure}

\subsection{Footstep Planning with the Robot's Full-Body Dynamics}
\label{sec:frbd_for_footstep_planning}
As explained in \sref{sec:related_work}, neither~\gls{srbd} nor~\gls{cd} can enforce joint torque limits.
Furthermore, ~\gls{srbd} cannot account for limb dynamics and kinematics.

Here, we present results that justify our full-body dynamics~\gls{mpc} that is able to consider the robot's joint torque limits and limb dynamics when planning footsteps placements and contact surfaces.
To strictly focus on the independent variables (i.e., joint torque limits and friction coefficients), we set up the simulation as shown in~\fref{fig:frbd_for_footstep_planning}.

\subsubsection{Joint Torque Limits}\label{sec:joint_torque_limits}
We investigated the ability of the robot to adjust to changing torque limits.
The result will allow us to consider having the robot carry a payload (e.g., an arm), or perform more dynamic movements (e.g., crossing large gaps), without exceeding the robot's torque limit.
First, we used the ANYmal's default torque limits of \SI{40}{\newton\meter} to have the robot walk with a trotting gait on the footstep surface of two pallets with a gap of \SI{30}{\centi\meter}.
Next, we reduced the torque limits by half.
\fref{fig:frbd_for_footstep_planning}a and \ref{fig:frbd_for_footstep_planning}b show the selection of footstep placements and contact surfaces for the default and reduced torque limits, respectively.
These demonstrate that our~\gls{mpc} with the reduced torque limit positions the robot's legs closer together.
As a result, the joint torque commands are reduced and kept within the torque limit, as shown in~\fref{fig:torque_plot}.

\begin{figure*}[htp]
    \centering\begin{tabular}{cc}
    \rowname{a} &
     {\raisebox{-.5\height}{\includegraphics[width=0.965\textwidth]{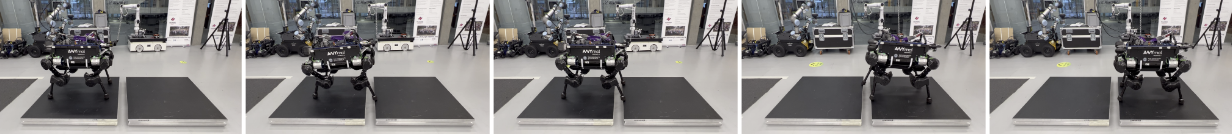}}}\\\\
    \rowname{b} & {\raisebox{-.5\height}{\includegraphics[width=0.965\textwidth]{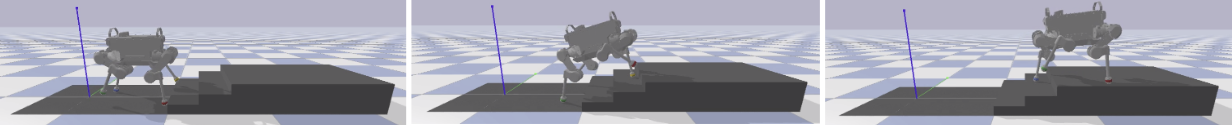}}}\\\\
    \rowname{c} & {\raisebox{-.5\height}{\includegraphics[width=0.965\textwidth]{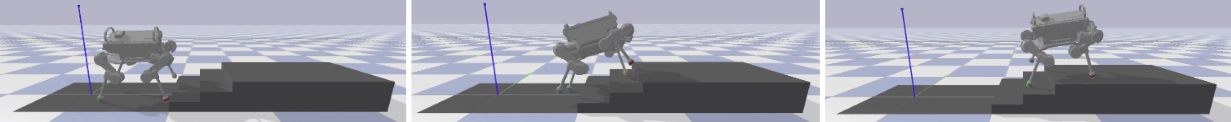}}}\\\\
    \rowname{d} & {\raisebox{-.5\height}{\includegraphics[width=0.965\textwidth]{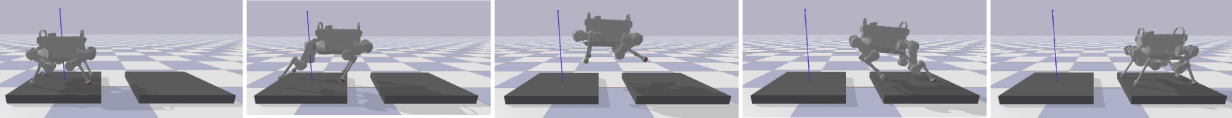}}}
    \end{tabular}
    \caption{
    Snapshots of various locomotion maneuvers computed by our topology-based~\gls{mpc} with automatic footstep placement and contact surface selection.
    All experimental and simulation trials were conducted using an~\gls{mpc} with an optimization horizon of \SI{0.85}{\second} and a control frequency of \SI{50}{\hertz}.
    To further explore the potential for more dynamic maneuvers that could be achieved with high-torque actuators, we increased the torque limit for simulations.
    (a) Experimental validation of a gap-crossing maneuver with a gap of~\SI{10}{\centi\meter}.
    (b) Climbing up a staircase of \SI{30}{\centi\meter} depth and \SI{10}{\centi\meter} height with a trotting gait.
    (c) Climbing up the same staircase with a pacing gait.
    (d) Jumping over a gap of \SI{40}{\centi\meter}.
    }
    \label{fig:robot_experiment}
\end{figure*}

\subsubsection{Limb Dynamics}
We varied the friction coefficient to investigate how the robot leverages limb dynamics when determining footstep placements and contact surfaces.
We used the ANYmal's default torque limits and the same reference velocity used in \sref{sec:joint_torque_limits}.
As shown in~\fref{fig:frbd_for_footstep_planning}c, our approach stretches the robot's leg perpendicular to the ground, allowing it to maintain the contact forces within the friction cone (with a friction coefficient of $0.14$).

\subsection{Validation of MPC on the ANYmal Robot}\label{sec:robot_experiment}
We validated our~\gls{mpc} approach on the ANYmal robot.  
The robot did not perceive its surroundings, but it received the candidate contact surface information extracted from the predefined environment, as shown in \fref{fig:control_pipeline}.
The robot moved with a walking gait.
It selected the next contact surface and planned its footstep placement within it, considering the full-body dynamics and kinematics, as shown in \fref{fig:robot_experiment}a.
With our topology-based~\gls{mpc}, the ANYmal robot successfully traversed a discrete terrain with a gap of \SI{10}{\centi\meter}.
This demonstrates that our approach can operate fast enough in an~\gls{mpc} scheme on real robots, with an optimization horizon of \SI{0.85}{\second} and a control frequency of \SI{50}{\hertz}.

\subsection{Achieving Dynamic Locomotion}\label{sec:potential_dynamic_motion_with_high_torque}
We conducted simulations to further evaluate our~\gls{mpc} scheme in achieving highly dynamic quadrupedal locomotion by utilizing limb dynamics.
To showcase the potential of our approach in generating highly dynamic maneuvers with robots equipped with high-torque actuators in the future, we disabled the robot's torque limits.

\subsubsection{Stair Climbing}
We extracted the contact surfaces of a staircase with \SI{30}{\centi\meter} depth and \SI{10}{\centi\meter} height.
Our~\gls{mpc} approach selected footstep placements and contact surfaces to climb stairs in real time.
The approach demonstrated the ability to handle both trotting and pacing dynamics, as shown in~\fref{fig:robot_experiment}b and~\ref{fig:robot_experiment}c.

\subsubsection{Dynamic Jumping}
We tested an even more dynamic jumping gait on terrain with a \SI{40}{\centi\meter} gap.
\fref{fig:robot_experiment}d shows that our~\gls{mpc} can achieve highly dynamic motions and adjust the footstep placements and contact surfaces to land reliably on the next contact surface.

% Add more detail about why this is an important validation result:
% 1. Steps are different convex patches
% 2. The proposed cost function is only 2D which means that when combined with the other cost functions it can be used on non-coplanar surfaces.
% 3. Steps are often occluded so replanning foot position based on newly detected contact surfaces provides more robustness.

%% file: src/5_conclusion.tex
\section{Conclusion}
In this work, we introduced a novel topology-based approach that enables full-body dynamics~\gls{mpc} to automatically select footstep placements and contact surfaces for locomotion over discrete terrains.
Specifically, we proposed a contact-surface penalty function that uses potential field and winding number to optimize both footstep placement and contact surface.
With our method, we first justified the importance of considering full-body dynamics, which includes joint torque limits and limb dynamics, in footstep planning.
We evaluated the planned footsteps and showed that our full-body dynamics~\gls{mpc} can effectively adapt to variations in joint torque limits and friction coefficients.
Second, to demonstrate the practical implementation of our method in an~\gls{mpc} scheme on real robots, we conducted hardware experiments on discrete terrain using the ANYmal quadruped robot.
Finally, we showcased the potential capabilities of our approach in various dynamic locomotion maneuvers with robots equipped with high-torque actuators through stair-climbing and gap-crossing simulations.

For future work, we plan to implement a perceptive locomotion pipeline using our~\gls{mpc} to demonstrate the practical usefulness of our approach in solving real-world problems with changing environments. Furthermore, we will conduct empirical comparisons between our full-body dynamics~\gls{mpc} approach and reduced-order dynamics~\gls{mpc} to highlight the significance of our proposed approach.
Lastly, as demonstrated in \fref{fig:footstep_planning}, we showed the potential of our method for humanoid locomotion. Although the control complexity is higher, our method can be considered a future research direction for humanoid footstep planning.